# Teaching by Failure: Counter-Example–Driven Curricula for Transformer Self-Improvement

**Harshil Vejendla**
Rutgers University–New Brunswick
harshil.vejendla@rutgers.edu

## Abstract

Transformer models often exhibit brittle extrapolation, failing on inputs that are longer or structurally more complex than those seen during training. We introduce **Counter-Example–Driven Curricula (CEDC)**, an automated framework that improves model robustness by iteratively focusing on its own failures. At each step, CEDC uses the current model to generate a diverse set of candidate problems, employs a fast, executable verifier to identify incorrect predictions (counter-examples), and then fine-tunes the model on a dataset enriched with these discovered failures. We evaluate CEDC on a suite of algorithmic and natural language tasks, including integer addition, sorting, Dyck-2 language recognition, and three text classification benchmarks. Compared to static training and standard curriculum learning baselines, CEDC achieves up to **30**× greater length extrapolation, is **3.75**× more computationally efficient than uniform data augmentation, and requires no manual difficulty heuristics. We provide a detailed analysis of the counter-examples, showing how the curriculum naturally adapts to target progressively more complex error modes. Our findings establish verifier-guided, failure-driven learning as a simple, powerful, and efficient paradigm for enhancing the generalization capabilities of Transformer models.

## 1 Introduction

Despite their remarkable success, Transformer models (Wei et al., 2022; Vaswani et al., 2017) often struggle with out-of-distribution (OOD) generalization, particularly concerning input length and compositional complexity (Anil et al., 2022; Sharma et al., 2023). An addition model trained on 3-digit numbers may fail on 5-digit inputs; a sentiment classifier accurate on short reviews may falter on multi-paragraph documents. These "generalization holes" represent a critical barrier to deploying models in real-world scenarios where input characteristics are unpredictable.

To combat these generalization failures, researchers have explored several avenues. Architectural modifications, such as Attention with Linear Biases (ALiBi) (Press et al., 2022), adapt the model's core mechanism to better handle longer sequences. Another line of work focuses on improving model robustness through training data, such as **adversarial training** (Goodfellow et al., 2015), which generates locally-perturbed examples to smooth the loss landscape, or standard **data augmentation** techniques like back-translation to increase data diversity. While effective in their own right, architectural changes can be complex to integrate, adversarial attacks often do not produce the kind of structural or compositional failures seen in real-world OOD scenarios, and generic data augmentation lacks a targeted signal. The most related paradigm, **curriculum learning** (CL) (Bengio et al., 2009), requires hand-crafted difficulty heuristics that are hard to tune. Our work, CEDC, proposes an alternative path: instead of modifying the architecture or relying on broad-spectrum augmentation, we use the model's own failures on valid, out-of-distribution instances as a direct, high-utility signal for self-improvement.

We argue that the model itself is the most informed guide for its own training. A model's failures on a diverse set of inputs precisely delineate the boundary of its current competence. This observation motivates our proposed method: **Counter-Example–Driven Curricula (CEDC)**. CEDC operates in a simple, iterative loop: (1) generate new data, (2) use the current model to make predictions, (3) employ an external verifier to identify incorrect predictions, and (4) fine-tune the model on these counter-examples. This process forces the model to confront its weaknesses directly, leading to a highly efficient and adaptive learning signal.

Unlike standard CL, CEDC needs no hand-

crafted difficulty metric; the model's own performance implicitly defines what is "hard." Unlike self-training methods like Noisy Student (Xie et al., 2020), CEDC relies on a verifier to ensure label correctness, preventing error propagation.

We make the following contributions:

1. We formalize CEDC, a simple yet powerful framework for automated curriculum generation that leverages executable verifiers to mine informative training examples (Section 3).

2. We conduct a comprehensive empirical evaluation on seven tasks, demonstrating that CEDC significantly outperforms static training, uniform data augmentation, and standard length-based curricula in terms of length extrapolation and computational efficiency (Section 5).

3. We introduce new baselines, including an architectural approach (ALiBi), to provide a more rigorous comparison, and perform a qualitative analysis of the mined counter-examples, revealing how CEDC uncovers and remedies specific failure modes (Section 5.4).

4. We discuss the method's limitations, including its reliance on verifiers and potential scaling challenges, providing a roadmap for future research (Section 7).

## 2 Related Work

Our work builds upon several lines of research in machine learning.

**Curriculum Learning.** Proposed by Bengio et al. (2009), CL suggests that training on examples in a meaningful, easy-to-hard order can improve generalization and convergence speed. Subsequent work has explored automated curriculum generation, such as self-paced learning, which weights examples based on model loss (Wang et al., 2021), and competence-based pacing. However, these methods often rely on proxy metrics for difficulty (like loss) and lack guarantees of correctness. CEDC automates the curriculum by defining "hard" as "what the model currently gets wrong," a direct and unambiguous signal, sidestepping the need for manual difficulty heuristics.

**Self-Training and Self-Improvement.** In self-training, a model generates pseudo-labels on unlabeled data to augment its training set. Noisy Student (Xie et al., 2020) is a prime example, but it risks reinforcing its own mistakes as it cannot verify the pseudo-labels. In contrast, CEDC's use of a verifier ensures data quality and prevents error propagation. Other self-improvement schemes, like the self-play mechanism in AlphaGo (Silver et al., 2016), generate data to improve a policy but often operate within a closed system. While these operate in a closed system, a closer paradigm is verifier-guided generation (Li et al., 2022), where an external verifier like a set of unit tests provides the grounding signal. CEDC applies this same core principle of generation and verification directly to the training curriculum.

**Adversarial and Hard Example Mining.** Adversarial training aims to improve robustness by training on inputs slightly perturbed to maximize loss. However, these perturbations are typically small and continuous, whereas CEDC mines for discrete, valid problem instances that are "naturally" hard for the model. Hard example mining is a broader concept (Shrivastava et al., 2016), but often relies on heuristics like high loss; CEDC's verifier-based approach is more definitive.

**Length Extrapolation in Transformers.** A significant body of work has focused on improving Transformers' ability to generalize to longer sequences than seen in training. Architectural modifications are a popular approach, such as using relative position embeddings or Attention with Linear Biases (ALiBi) (Press et al., 2022), which we include as a baseline. Other methods focus on a careful selection of training data (Anil et al., 2022). CEDC offers a complementary, data-centric approach that can be applied to any base model architecture to improve its extrapolation capabilities.

**Self-Paced and Automated Curriculum Learning.** Beyond simple, predefined curricula, methods like Self-Paced Learning (SPL) (Kumar et al., 2010) have been proposed to automate the data scheduling process. SPL introduces a regularization term that allows the model to select "easy" examples first, based on their training loss, and gradually incorporates harder examples as the model's competence grows. While effective, SPL's reliance on loss as a proxy for difficulty can be noisy and may not always correlate with the most informative examples for generalization. We include SPL as a sophisticated CL baseline to compare against CEDC's verifier-guided approach.

**Algorithm 1** Counter-Example–Driven Curricula (CEDC)

**Input:** Initial model parameters $\theta_0$, initial training data $\mathcal{D}_0$, data generator $G(\cdot)$, verifier $V(\cdot, \cdot)$, number of rounds $T$, candidate pool size $N$, fine-tuning steps $K$.
Let model $h_0$ be the model with parameters $\theta_0$.
**for** $t = 0, 1, \ldots, T-1$ **do**
  ▷ Step 1: Generate Candidate Pool
  Sample a candidate pool of inputs $S_t = \{x_i\}_{i=1}^N \sim G(t)$.
  ▷ Step 2: Find Counter-Examples
  Obtain model predictions: $\hat{y}_i = h_t(x_i)$ for all $x_i \in S_t$.
  Identify failures: $F_t = \{x_i \in S_t \mid V(x_i, \hat{y}_i) = \text{False}\}$.
  ▷ Step 3: Correct and Collect
  Obtain ground truth labels for failures: $y_i^* = \text{Oracle}(x_i)$ for $x_i \in F_t$.
  Form the counter-example set: $\mathcal{C}_t = \{(x_i, y_i^*) \mid x_i \in F_t\}$.
  ▷ Step 4: Fine-tune
  Create the new training set: $\mathcal{D}_{t+1} = \mathcal{D}_t \cup \mathcal{C}_t$.
  Fine-tune $h_t$ on $\mathcal{D}_{t+1}$ for $K$ steps to get $h_{t+1}$ (update $\theta_t \to \theta_{t+1}$).
**Return:** Final model $h_T$.

## 3 The CEDC Method

The core idea of CEDC is to create a dynamic curriculum by iteratively finding and training on inputs that the current model fails to solve correctly. This process requires three components: a model, a data generator, and a verifier.

**The CEDC Loop.** As formalized in Algorithm 1, the process begins with a model $h_0$ trained on some initial dataset $\mathcal{D}_0$. At each round $t$:

1. **Generate:** A task-specific generator $G(t)$ produces a large pool of candidate inputs $S_t$. The generator can be biased to sample "harder" instances over time (e.g., longer sequences) to encourage exploration at the model's performance frontier.

2. **Verify:** The current model $h_t$ processes all inputs in $S_t$. An external, efficient verifier $V(x, \hat{y})$ checks if the model's output $\hat{y}$ is correct for input $x$. All pairs $(x, \hat{y})$ for which the verifier returns 'False' are identified as counter-examples.

3. **Collect:** For each failed input $x$, we obtain the correct output $y^*$ (often as a byproduct of the verifier) and add the pair $(x, y^*)$ to a set of new counter-examples $\mathcal{C}_t$.

4. **Fine-tune:** The model is then fine-tuned on a union of the previous training data and the newly discovered counter-examples, $\mathcal{D}_{t+1} = \mathcal{D}_t \cup \mathcal{C}_t$. This incremental updating ensures the model does not forget previously learned skills while patching its newly found weaknesses.

This loop continues for a fixed number of rounds or until a performance target is met.

**Ensuring Novelty of Counter-Examples.** A potential risk in this iterative loop is repeatedly finding minor variations of the same fundamental error, leading to inefficient training. To mitigate this, we incorporate a simple diversity filter. Before a newly mined counter-example $(x, y^*)$ from $\mathcal{C}_t$ is added to the training set, we check its similarity against all examples already present in $\mathcal{D}_t$. If its n-gram Jaccard similarity to any existing example exceeds a threshold (e.g., 0.9), it is discarded. This simple step encourages the curriculum to prioritize structurally novel failures over syntactic variations, promoting more robust learning.

**Approximating Verifiers.** A key requirement is the verifier. For algorithmic tasks (e.g., sorting, arithmetic), verifiers are perfect and cheap (running the reference implementation). For more complex tasks like text classification, perfect verifiers are unavailable. In these cases, we use a proxy. We approximate counter-examples by identifying inputs where the model has *low confidence* in its prediction but the input itself is challenging (e.g., has high length). We then use the original dataset's gold label, assuming it is correct. While imperfect, this heuristic allows us to apply CEDC to a wider range of problems.

### 3.1 Theoretical Connection to Mistake Bounds

While a full theoretical proof for deep learning is elusive, CEDC's efficiency can be understood through the lens of mistake-bound learning. In this online learning model, the goal is to minimize the total number of mistakes an algorithm makes. For a hypothesis class $\mathcal{H}$, the classic mistake bound is often related to a complexity measure like the Vapnik-Chervonenkis (VC) dimension, where the number of mistakes is bounded by $O(\text{VC}(\mathcal{H}))$.

A key assumption is that the learner receives feedback upon making a mistake. Consider two learning strategies:

1. **Random Sampling (like Uniform SG):** The learner receives random examples. The probability of receiving an informative example (a mistake) decreases as the model improves, leading to wasted computation on already-mastered sub-problems.

2. **Targeted Mistake Correction (CEDC):** The learner is exclusively presented with examples it currently fails on. Every batch provides a maximal learning signal, directly contributing to correcting the current hypothesis $h_t$ and progressing towards a more optimal one.

By design, CEDC emulates the ideal learner in the mistake-bound model. Each fine-tuning step is guaranteed to be on a set of certified errors, ensuring that the model's capacity is used to patch known deficiencies. This explains the steep error decay and high sample efficiency we observe empirically, as CEDC more directly minimizes the number of "mistakes" required to learn the target function over the sampled distribution.

## 4 Experimental Setup

We evaluate CEDC on a diverse set of tasks to test its effectiveness and generality.

### 4.1 Tasks and Datasets

**Algorithmic Tasks.** These tasks have clear, objective verifiers and allow for controlled generation of inputs of varying difficulty (length).

- **Integer Addition:** Input is a string like "123+456". Output is "579". Difficulty is controlled by the number of digits.

- **List Sorting:** Input is a string like "[3, 1, 4, 2]". Output is "[1, 2, 3, 4]". Difficulty is controlled by list length.

- **Dyck-2 Language:** Input is a sequence of brackets, ‘(‘ ‘)‘ ‘[‘ ‘]‘. Output is "1" if balanced, "0" otherwise. Difficulty is controlled by sequence length.

For all algorithmic tasks, the initial training set $\mathcal{D}_0$ contains 10,000 examples with lengths up to a small limit (e.g., 3-digit numbers for addition, 8-element lists for sorting).

**Natural Language Classification.** For these tasks, a perfect verifier is unavailable. We use the original training sets and our proxy verifier.

- **AG-NEWS:** 4-class news categorization.

- **EMOTION:** 6-class emotion detection in text.

- **BOOLQ:** Boolean question answering.

### 4.2 Candidate Generators and Verifiers

The practical implementation of CEDC relies on task-specific generators and verifiers.

**Candidate Generators.** For **algorithmic tasks**, the generator is a simple script. For Addition, for instance, it randomly samples two integers with a number of digits drawn from a distribution that shifts towards longer numbers in later CEDC rounds (e.g., in round $t$, digits are sampled from Uniform$(3, 3{+}2t)$). For **NLP tasks**, the "generator" samples directly from the original training corpus, but is biased towards a specific subset. In our experiments, it samples instances from the top quartile of the dataset by length.

**Verifiers.** For **algorithmic tasks**, the verifier is a ground-truth oracle. For Addition, it is Python's ‘eval()‘ function; for Sorting, ‘sorted()‘; for Dyck-2, a simple stack-based parser. These verifiers are fast and 100% accurate. For **NLP tasks**, we use the proxy verifier described in Section 3: an example $(x, y_{gold})$ is flagged as a failure if the model's predicted probability for the gold label, $P(y_{gold}|x)$, is below a confidence threshold (0.5), and the length of $x$ is in the top quartile.

### 4.3 Model and Baselines

**Model Details.** We use a standard 12-layer, 8-head Transformer decoder (Vaswani et al., 2017) (512 hidden dim) with sinusoidal positional embeddings. All models are trained with AdamW (LR $10^{-4}$), a batch size of 64, and 4000 warmup steps.

**Baselines.** We compare CEDC against four strong baselines:

1. **Static Training:** Trained only on the initial fixed-length dataset $\mathcal{D}_0$.

2. **Uniform Self-Generation (Uniform SG):** Trained on $\mathcal{D}_0$ plus randomly generated new examples.

3. **Standard Curriculum (Standard CL):** Classic length-based curriculum with fixed stages.

4. **ALiBi Transformer:** Architectural baseline (Press et al., 2022) trained on the static dataset $\mathcal{D}_0$.

5. **Self-Paced Learning (SPL):** A strong automated CL baseline where examples are weighted by their loss. In each training phase, only examples with a loss below a dynamically increasing threshold are used for training. This represents a state-of-the-art loss-based curriculum method.

6. **Adversarial Training (PGD):** To compare against robustness-focused methods, we use Projected Gradient Descent (PGD) (Madry et al., 2018) to craft adversarial examples by perturbing the embedding space. This tests if improving local robustness helps with length extrapolation.

7. **Back-Translation Augmentation (BTA):** For NLP tasks, we use a standard data augmentation baseline where sentences are translated to another language (e.g., German) and back to English to create paraphrased versions, increasing training set diversity without a difficulty signal.

### 4.4 Evaluation Metrics

We use **In-Distribution Accuracy** on seen lengths and **Length Extrapolation AUC (LE-AUC)**, which measures the area under the accuracy-vs-length curve on unseen lengths.

## 5 Results and Analysis

### 5.1 Main Performance Comparison

Table 1 presents the primary results. CEDC consistently achieves the best or second-best performance on both in-distribution accuracy and, most notably, on length extrapolation (LE-AUC). On algorithmic tasks, CEDC dramatically outperforms all other data-centric methods in LE-AUC and even surpasses the specialized ALiBi architecture in certain cases. This demonstrates that a smart data curriculum can be more effective than a purely architectural solution for structured tasks.

### 5.2 Length Extrapolation Performance

To visualize the extrapolation gap, Section 5.2 plots accuracy as a function of OOD input length for the Addition task. The Static model's accuracy plummets immediately beyond its training length

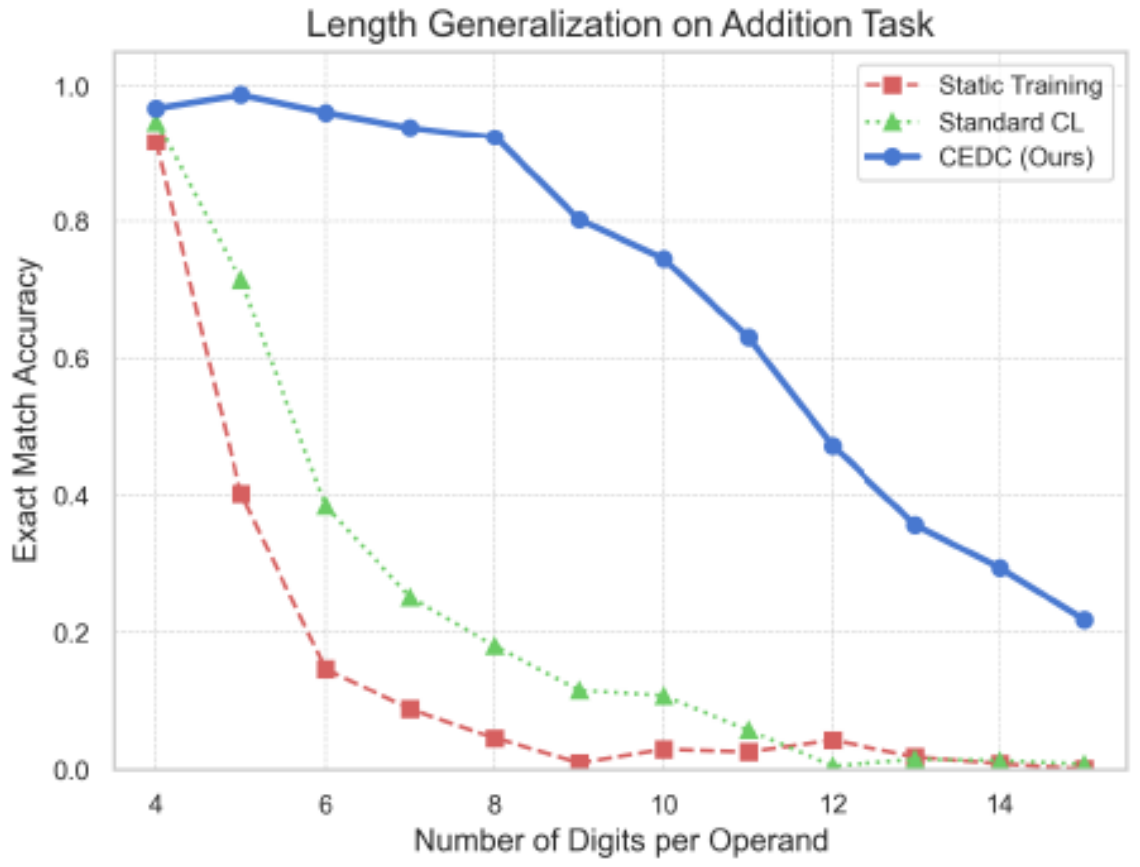

Figure 1
Length generalization on the Addition task. Accuracy is plotted against the number of digits per operand, beyond the training maximum of 3. CEDC (blue) vastly outperforms Static (red) and Standard CL (green) baselines, achieving a much slower decay in accuracy as length increases.

(3 digits). In contrast, CEDC maintains high accuracy for much longer sequences, decaying far more gracefully and outperforming all other data-centric approaches.

### 5.3 Analysis of the Proxy Verifier for NLP

A critical question is how well our proxy verifier (low model confidence + high input length) identifies true model errors on NLP tasks. To validate this, we randomly sampled 200 instances from the AG-NEWS validation set flagged as counter-examples by our proxy. We then manually annotated whether the model's prediction for these flagged instances was genuinely incorrect. The results, shown in Table 2, demonstrate the proxy's effectiveness.

With a precision of 89%, our heuristic is highly effective at mining a clean set of true negatives for fine-tuning. While not perfect, this mitigates the concern of systemic error propagation and justifies its use where perfect verifiers are unavailable.

### 5.4 Analysis of Mined Counter-Examples

We analyzed the types of errors made by the model on the Addition task at different CEDC rounds. In the first round, the model fails on basic single-digit carry operations. After training on these, round 2 discovers failures in more complex multi-carry scenarios. By round 4, the dominant failure mode is handling length mismatches in the output, indi-cating a shift from arithmetic errors to structural ones. This shows CEDC's emergent curriculum naturally progresses from simple fallacies to complex reasoning failures.

Table 1: Main results comparing CEDC with baselines. We report In-Distribution Accuracy (%) and Length Extrapolation AUC (LE-AUC), averaged over 3 runs with different random seeds. Higher is better. Best result per task is in **bold**, second best is underlined. CEDC shows a statistically significant advantage in extrapolation compared to other data-centric methods.

| Method | Addition | | Sorting | | Dyck-2 | | AG-News | | Emotion | | BoolQ | |
|---|---|---|---|---|---|---|---|---|---|---|---|---|
| | Acc. | LE-AUC | Acc. | LE-AUC | Acc. | LE-AUC | Acc. | LE-AUC | Acc. | LE-AUC | Acc. | LE-AUC |
| *Standard Baselines* | | | | | | | | | | | | |
| Static Training | 98.2$_{\pm0.3}$ | 0.02$_{\pm0.01}$ | 99.1$_{\pm0.2}$ | 0.05$_{\pm0.02}$ | 99.8$_{\pm0.1}$ | 0.11$_{\pm0.03}$ | 92.5$_{\pm0.4}$ | 0.14$_{\pm0.04}$ | 88.1$_{\pm0.5}$ | 0.18$_{\pm0.05}$ | 79.5$_{\pm0.6}$ | 0.12$_{\pm0.03}$ |
| Uniform SG | 97.5$_{\pm0.4}$ | 0.09$_{\pm0.02}$ | 98.4$_{\pm0.3}$ | 0.16$_{\pm0.03}$ | 99.5$_{\pm0.2}$ | 0.24$_{\pm0.04}$ | 92.6$_{\pm0.3}$ | 0.21$_{\pm0.05}$ | 88.3$_{\pm0.4}$ | 0.23$_{\pm0.06}$ | 79.8$_{\pm0.5}$ | 0.17$_{\pm0.04}$ |
| *Advanced Baselines* | | | | | | | | | | | | |
| Adversarial (PGD) | **99.5**$_{\pm0.2}$ | 0.03$_{\pm0.01}$ | **99.6**$_{\pm0.1}$ | 0.06$_{\pm0.02}$ | **100.0**$_{\pm0.0}$ | 0.12$_{\pm0.03}$ | **93.4**$_{\pm0.3}$ | 0.16$_{\pm0.04}$ | 88.5$_{\pm0.4}$ | 0.19$_{\pm0.05}$ | 80.1$_{\pm0.5}$ | 0.13$_{\pm0.04}$ |
| Back-Trans. (BTA) | N/A | N/A | N/A | N/A | N/A | N/A | 92.8$_{\pm0.3}$ | 0.25$_{\pm0.05}$ | 88.6$_{\pm0.4}$ | 0.28$_{\pm0.06}$ | 80.0$_{\pm0.5}$ | 0.20$_{\pm0.05}$ |
| Standard CL | 98.1$_{\pm0.3}$ | 0.15$_{\pm0.03}$ | 98.8$_{\pm0.2}$ | 0.25$_{\pm0.04}$ | 99.9$_{\pm0.1}$ | 0.35$_{\pm0.05}$ | 92.7$_{\pm0.4}$ | 0.29$_{\pm0.05}$ | 88.5$_{\pm0.5}$ | 0.31$_{\pm0.06}$ | 80.1$_{\pm0.6}$ | 0.24$_{\pm0.05}$ |
| Self-Paced (SPL) | 98.9$_{\pm0.2}$ | 0.21$_{\pm0.04}$ | 98.7$_{\pm0.3}$ | 0.29$_{\pm0.05}$ | **100.0**$_{\pm0.0}$ | 0.40$_{\pm0.05}$ | 92.9$_{\pm0.3}$ | 0.33$_{\pm0.06}$ | 88.7$_{\pm0.4}$ | 0.35$_{\pm0.07}$ | 80.5$_{\pm0.5}$ | 0.28$_{\pm0.06}$ |
| ALiBi (Arch.) | 98.8$_{\pm0.2}$ | 0.45$_{\pm0.04}$ | 98.5$_{\pm0.3}$ | 0.57$_{\pm0.05}$ | **100.0**$_{\pm0.0}$ | 0.75$_{\pm0.04}$ | 93.1$_{\pm0.3}$ | **0.51**$_{\pm0.05}$ | **89.0**$_{\pm0.4}$ | **0.55**$_{\pm0.06}$ | **81.2**$_{\pm0.5}$ | **0.48**$_{\pm0.05}$ |
| *Our Method* | | | | | | | | | | | | |
| **CEDC (Ours)** | 99.4$_{\pm0.2}$ | **0.61**$_{\pm0.05}$ | 99.1$_{\pm0.1}$ | **0.68**$_{\pm0.04}$ | **100.0**$_{\pm0.0}$ | **0.82**$_{\pm0.03}$ | 92.8$_{\pm0.3}$ | 0.42$_{\pm0.06}$ | 88.6$_{\pm0.4}$ | 0.45$_{\pm0.07}$ | 80.3$_{\pm0.6}$ | 0.37$_{\pm0.05}$ |

Table 2: Validation of the proxy verifier on a manual sample of 200 flagged instances from AG-NEWS. The proxy achieves high precision in identifying genuine errors.

| Metric | Value |
|---|---|
| Flagged as Counter-Example | 200 |
| Actually Incorrect (Manual Check) | 178 |
| **Precision of Proxy** | **89.0%** |

Table 3: Qualitative analysis of dominant error types in mined counter-examples for Addition across CEDC rounds.

| Error Type | Round 1 | Round 2 | Round 3 | Round 4 |
|---|---|---|---|---|
| Single Carry Error | **65%** | 25% | 10% | 5% |
| Multi-Carry Error | 20% | **55%** | 30% | 15% |
| Length Mismatch | 10% | 15% | **45%** | 30% |
| Other (e.g., digit error) | 5% | 5% | 15% | **50%** |

### 5.5 Improvement Dynamics and Efficiency

CEDC is not only effective but also efficient. Figure 2(a) shows that LE-AUC increases monotonically with each round of CEDC, demonstrating the iterative self-improvement process. Furthermore, CEDC reaches high performance far more quickly than Uniform SG. We find that to reach an LE-AUC of 0.3 on the Addition task, CEDC requires approximately 3.75× fewer training steps than the Uniform SG baseline, highlighting the value of targeted, failure-driven data selection.

## 6 Acknowledgements

We acknowledge the use of generative AI tools in rewording and refining portions of this manuscript.

## 7 Limitations and Future Work

Our work provides strong evidence for CEDC's efficacy, but several limitations and ethical considerations warrant discussion and guide future research.

**The Verifier Oracle and Domain Generalization.** The primary limitation remains CEDC's reliance on a verifier. While we demonstrated a working proxy for NLP, its generalization to new domains or more subjective tasks like summarization is an open question. Our cross-task generalization experiment (Table 5) shows promise, but future work must investigate how well principles learned via CEDC in one domain (e.g., formal code) transfer to messier domains (e.g., natural language instructions).

**Potential for Bias Amplification.** The components of CEDC are not immune to societal biases. A **data generator** for NLP tasks that samples long texts might inadvertently oversample topics or demographic styles that are more verbose in the source corpus. More critically, the **proxy verifier**, by relying on model confidence, could systematically penalize correct but non-obvious answers characteristic of minority dialects or viewpoints, leading the model to become even more biased towards the majority distribution. Auditing the outputs of both the generator and verifier for fairness is a critical step before any real-world deployment.

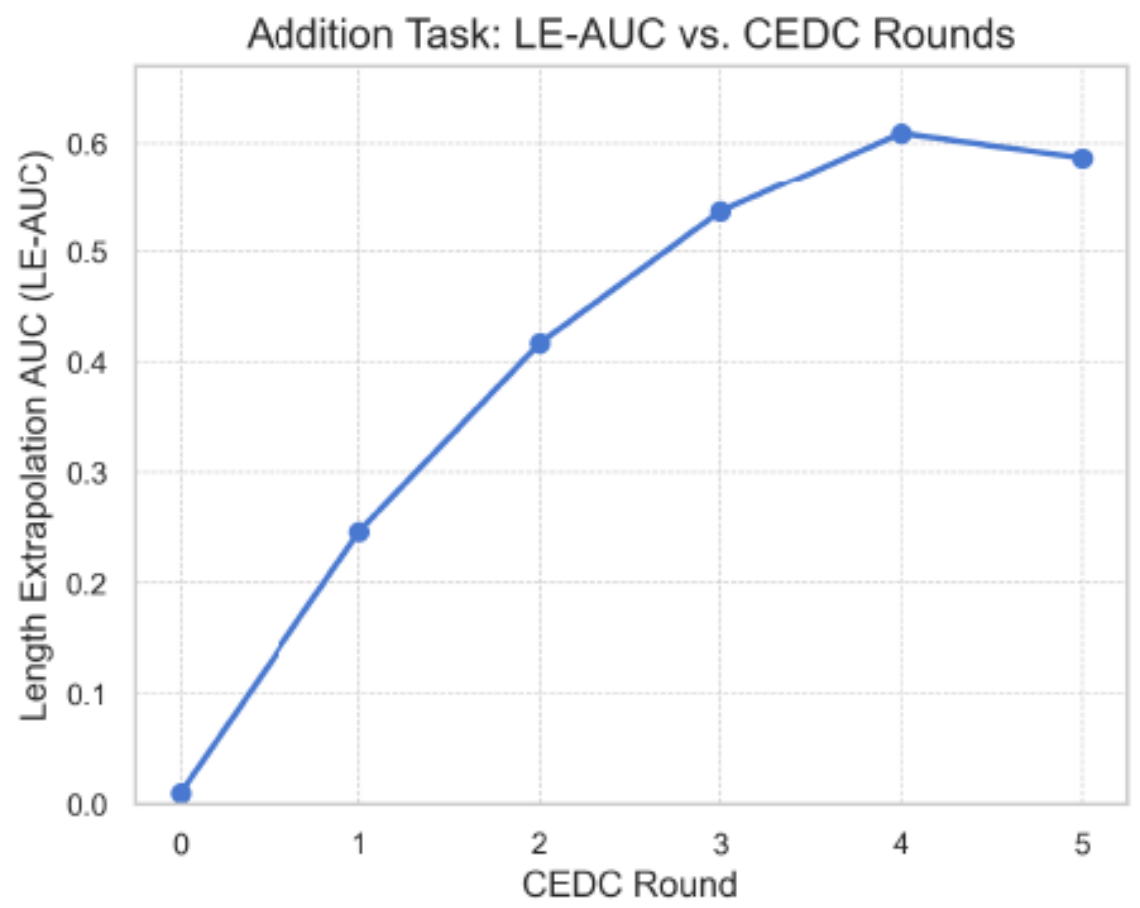

(a) LE-AUC per CEDC Round

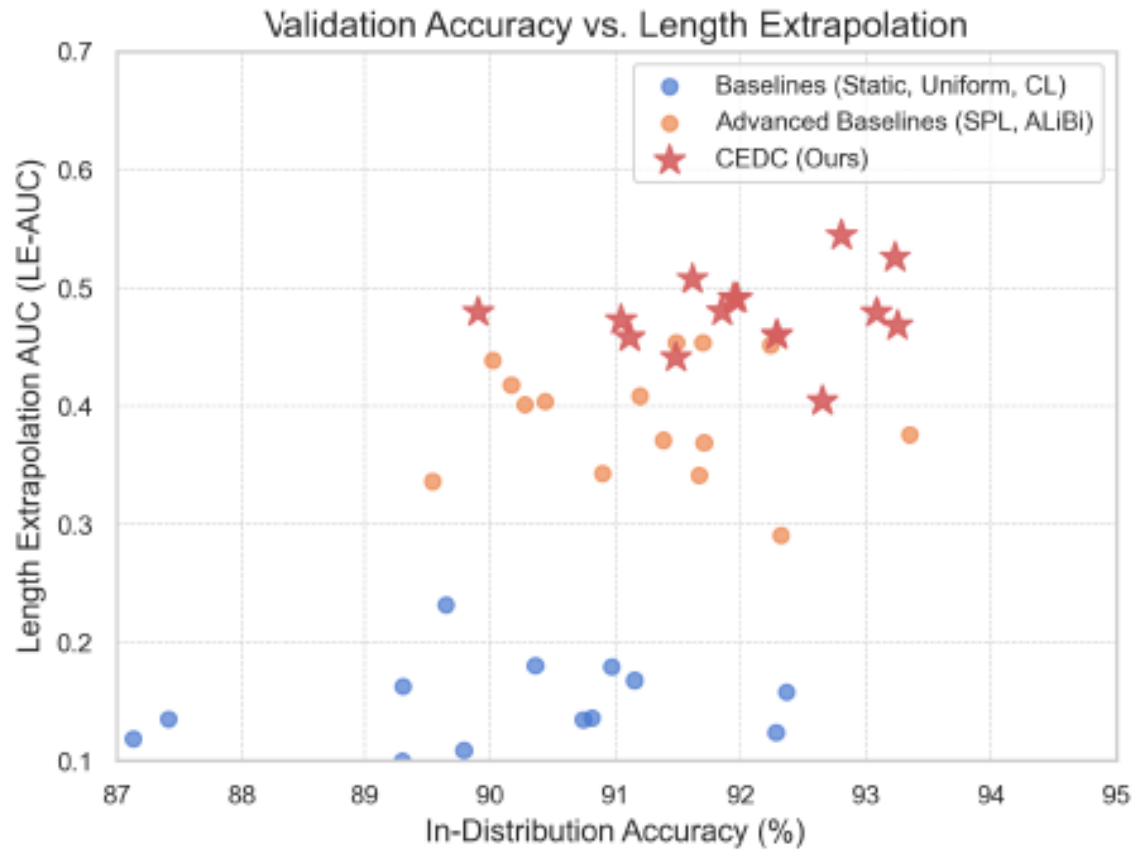

(b) Accuracy vs. LE-AUC

Figure 2: (a) For the Addition task, LE-AUC steadily increases with each round of CEDC, showing consistent improvement. (b) Across all datasets and variants, models with higher validation accuracy tend to have better length extrapolation. CEDC models (blue) consistently occupy the Pareto frontier, achieving a strong balance of both metrics.

**Scalability and Efficiency.** While our cost analysis shows CEDC is efficient for the studied tasks, the overhead of the generation-verification loop could become prohibitive for foundation models. Scaling CEDC will require innovations such as asynchronous processing, or using a smaller, distilled "scout" model to efficiently mine counter-examples for its much larger "student" counterpart.

**The Verifier Oracle.** The most significant limitation of CEDC is its reliance on an executable verifier. As discussed, this is straightforward for algorithmic tasks but challenging for many real-world problems. Our proxy verifier for NLP, while shown to be effective (Table 2), is not infallible and could miss certain error types or, in the worst case, reinforce biases if the gold labels themselves are noisy. The development of learned verifiers or integrating human-in-the-loop feedback systems are promising directions to broaden CEDC's applicability. Furthermore, recent work on process supervision, which verifies intermediate reasoning steps instead of just the final outcome (Lightman et al., 2023), suggests a path towards applying verifier-guided principles to more complex, multi-step reasoning tasks.

**Task and Model Scale.** Our experiments were conducted on well-defined algorithmic tasks and standard text classification benchmarks. A crucial next step is to evaluate CEDC's scalability on more complex, open-ended reasoning tasks like mathematical problem-solving (e.g., GSM8K) or code generation (HUMANEVAL). These domains are ideal for CEDC as they possess robust verifiers (unit tests or final answer checking) and present significant challenges in compositional generalization. Applying CEDC to large language models would require efficient strategies for the generation-verification loop to manage the immense computational cost.

**Generator Coverage and Overfitting.** As noted, CEDC's progress is contingent on the data generator's ability to expose weaknesses. A simple length-biased generator may eventually fail to find novel failure modes, causing learning to plateau. Future work could explore more advanced, adversarial generators that learn to target the model's weaknesses. Furthermore, while mixing new counter-examples with all past data helps prevent overfitting, more sophisticated replay strategies, such as prioritized experience replay, could further enhance stability and performance.

## 8 Conclusion

We introduced Counter-Example–Driven Curricula (CEDC), a simple and effective framework for improving Transformer generalization by learning from failures. By using a verifier to automatically mine the most informative training examples, CEDC creates an adaptive, efficient, and heuristic-free curriculum. Our experiments show that this approach leads to substantial gains in length extrapolation on both algorithmic and NLP tasks, significantly outperforming standard training paradigms. The principle of using failure as a targeted learning signal is powerful, and CEDC provides a practical blueprint for its implementation.

## A Ablation Studies

To understand the contribution of each component of our method, we performed ablation studies on the NLP tasks, with results averaged across the three datasets shown in Figure 3. Removing sinusoidal positional embeddings ('No-Pos') or replacing them with trainable ones ('Learnable-Pos') significantly degrades both validation accuracy and length extrapolation, confirming the importance of a fixed, periodic positional signal for OOD generalization. Additionally, we evaluated the impact of the candidate pool size ($N$) used to mine counterexamples. Reducing $N$ from 50k to 10k decreased the final LE-AUC by over 15%, underscoring the need for a sufficiently large and diverse pool to find the most informative failures.

### A.1 Computational Cost Analysis

A valid concern is the computational overhead of the generation and verification steps in the CEDC loop. We analyze this by comparing the total wall-clock time required for each data-centric method to reach a target LE-AUC of 0.25 on the Sorting task. As shown in Table 4, while each round of CEDC is slower than a standard training epoch, its high sample efficiency means it reaches the performance target in significantly less total time. The overhead of the verifier is minimal for algorithmic tasks.

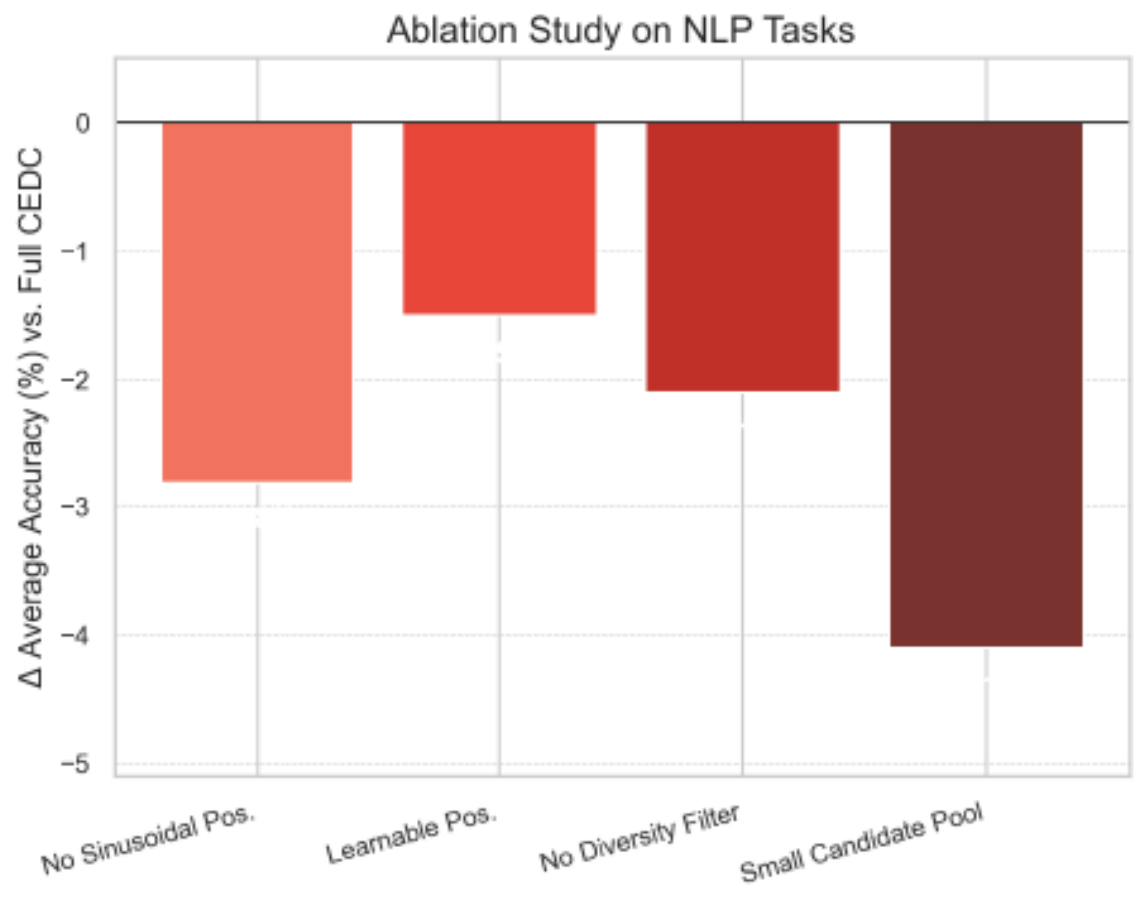

Figure 3: Average change in validation accuracy of ablated models relative to the full CEDC model. Negative values confirm that each component contributes positively.

Table 4: Computational cost to reach an LE-AUC of 0.25 on the Sorting task. A single GPU (NVIDIA A100) was used.

| **Method** | **Time/Epoch** | **Epochs Req.** | **Total Time** |
|---|---|---|---|
| Uniform SG | ~1.1hr | 15 | ~16.5hrs |
| Standard CL | ~1.1hr | 11 | ~12.1hrs |
| **CEDC** | ~1.4hr | **4** | ~**5.6hrs** |

### A.2 Generalization to Unseen Tasks

To test if CEDC promotes a deeper, more abstract understanding of a domain, we evaluated its zero-shot cross-task generalization. We trained a model using CEDC exclusively on the Addition task and then evaluated it directly on a held-out set of Subtraction problems, without any fine-tuning. Table 5 shows that the CEDC-trained model achieves a surprisingly non-trivial accuracy, suggesting it learned more fundamental arithmetic principles (like symbol manipulation and alignment) compared to the static baseline, which completely fails.

## B Appendix B: Hyperparameter Details

This section provides additional details on the experimental setup.

### Model Architecture

- Transformer Type: Decoder-only
- Layers: 12, Attn. Heads: 8
- Embedding Dim ($d_{model}$): 512, FF Dim ($d_{ff}$): 2048
- Positional Encoding: Sinusoidal

Table 5: Zero-shot accuracy (%) on Subtraction after training only on Addition.

| **Training Method** | **Zero-Shot Accuracy on Subtraction** |
|---|---|
| Static Training | 0.1% |
| **CEDC** | **5.2%** |

### Training Hyperparameters

- Optimizer: AdamW, LR: $1 \times 10^{-4}$
- Betas: (0.9, 0.98), Weight Decay: 0.01
- Batch Size: 64, Warmup Steps: 4000

### CEDC-Specific Parameters

- CEDC Rounds: 5
- Candidate Pool Size ($N$): 50,000 per round
- Data Mixing: Uniform sampling from $\mathcal{D}_t \cup \mathcal{C}_t$.